\documentclass[a4paper, 10 pt, conference]{ieeeconf}
\IEEEoverridecommandlockouts 
\usepackage[utf8]{inputenc} 
\usepackage[T1]{fontenc}    
\usepackage{hyperref}       
\usepackage{url}            
\usepackage{booktabs}       
\usepackage{amsfonts}       
\usepackage{nicefrac}       
\usepackage{microtype}      
\usepackage{lipsum}
\usepackage[T1]{fontenc}

\usepackage{cite}
\usepackage{balance}
\usepackage{amsmath,amssymb}
\usepackage{algorithmic}
\usepackage{graphicx}
\usepackage{subfigure}
\usepackage{textcomp}
\usepackage{gensymb}
\usepackage{epsfig} 
\usepackage{xcolor}

\usepackage{booktabs}
\usepackage{array}
\usepackage{multirow}

\usepackage[font=small,labelfont=bf]{caption} 

\title{Getting a Grip: {\em in Materio} Evolution of Membrane Morphology for Soft Robotic Jamming Grippers}

\author{David Howard$^{1}$, Jack O'Connor$^{1,2}$, Jordan Letchford$^{1,3}$, James Brett$^{1}$, Therese Joseph$^{1,3}$, \\Sophia Lin$^{1,4}$, Daniel Furby$^{1,3}$, and Gary W. Delaney$^{1}$
\thanks{$^{1}$ CSIRO, Australia; contact {david.howard@csiro.au}}
\thanks{$^{2}$ University of Queensland, Australia}
\thanks{$^{3}$ Queensland University of Technology, Australia}
\thanks{$^{4}$ University of Melbourne, Australia}
}

\begin{document}
\maketitle
\thispagestyle{empty}
\pagestyle{empty}

\begin{abstract}

The application of granular jamming in soft robotics is a recent and promising new technology offer exciting possibilities for creating higher performance robotic devices. Granular jamming is achieved via the application of a vacuum pressure inside a membrane containing particulate matter, and is particularly interesting from a design perspective, as a myriad of design parameters can potentially be exploited to induce a diverse variety of useful behaviours.  To date, the effect of variables such as grain shape and size, as well as membrane material, have been studied as a means of inducing bespoke gripping performance, however the other main contributing factor, membrane morphology, has not been studied due to its particular complexities in both accurate modelling and fabrication.  This research presents the first study that optimises membrane morphology for granular jamming grippers, combining multi-material 3D printing and an evolutionary algorithm to search through a varied morphology design space {\em in materio}.  Entire generations are printed in a single run and gripper retention force is tested and used as a fitness measure.  Our approach is relatively scalable, circumvents the need for modelling, and guarantees the real-world performance of the grippers considered.  Results show that membrane morphology is a key determinant of gripper performance.  Common high performance designs are seen to optimise all three of the main identified mechanisms by which granular grippers generate grip force, are significantly different from a standard gripper morphology, and generalise well across a range of test objects. 

\end{abstract}


\section{Introduction}

Granular Jamming \cite{fitzgerald_review_2020} is a popular and versatile soft robotic mechanism allowing high stiffness variation with minimal volume variation. By far the most prevalent use of granular jamming in the literature is the ‘Universal Gripper’, which consists of a spherical elastomeric balloon filled with coffee grounds, which is attached to a vacuum pump.  Applying a vacuum force causes the coffee to increase rigidity and grip onto a target object. 

Design of granular jamming grippers is an interesting and challenging research problem.  {\em Interesting}, because granular jamming structures are relatively unconstrained in their possible shapes and sizes.  Moreover, numerous design variables (including shape, size, and constituent materials of both grains and membranes \cite{fitzgerald_evolving_2021}), can potentially be tuned to elicit high gripper performance.  {\em Challenging} due to strongly-coupled and complex interactions between grains, membrane, and environment that are particularly difficult to model. Granular grippers are therefore largely unsuitable for model-based optimisation and design explorations of granular grippers have to date been limited in scope.

\begin{figure}[t!]
\centering
\includegraphics[width=0.9\columnwidth]{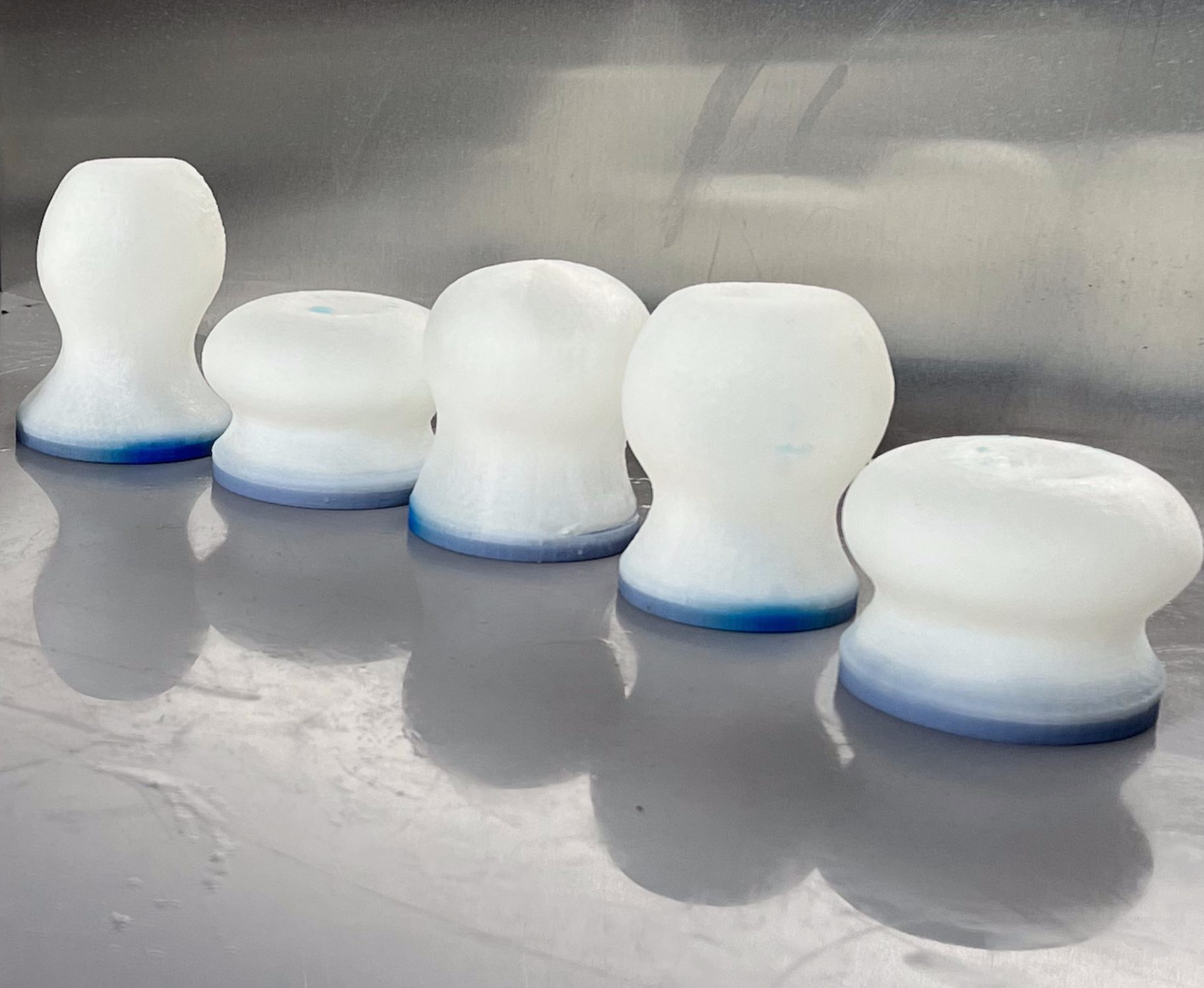}
\caption{A selection of evolved membranes, after printing and cleaning but pre-filling.}
\label{fig1}
\end{figure}

This is particularly true for membrane morphology, where complex deformations are tricky to capture via modelling and prone to reality gap effects, but may be a precursor to novel,  high performance designs \cite{howard_one-shot_2021}.  As such, the most common membrane over the past decade remains a simple spherical latex balloon.

We present the first study that optimises membrane morphology for granular jamming grippers, and the first study that directly optimises membrane morphology for a soft gripper in general.  It is particularly suited for soft robot design spaces that are not easily amenable to modelling, such as granular jamming, and encourages design methodologies based on embodied cognition, which hinge on high-fidelity interactions between an agent's constituent parts and an environment.

Our model-free optimisation technique combines 3D printing and a Genetic Algorithm (GA) to iteratively evolve populations of grippers {\em in materio} \cite{greenwood2006introduction}:  each generation is printed in one run on a multi-material 3D printer, before being post-processed and tested. No modelling is required. Additionally, the 3D printer allows for large theoretical maximum population sizes of up to 80 grippers to be printed at once, making the approach relatively scaleable.  Our approach also partially closes the loop on this design exploration via scripting: printable CAD models are automatically generated from gripper genomes, and fitness values are automatically returned from the experimental setup to the genetic algorithm. 

To demonstrate the feasibility of this approach, we evolve a total of 75 grippers across 15 generations.  Grippers are experimentally assessed for their ability to grip a challenging benchmark object.  Results show strong optimisation of retention force.   Common high performance designs are seen to optimise all three of the main identified mechanisms by which granular grippers generate grip force.  A wide range of geometries are generated, showing that the design space made available by the Bezier representation is useful and permits diverse designs, as well as suggesting the ability to adapt morphologies for diverse applications outside of gripping. (Fig.\ref{fig1}).  The best gripper performs well across a range of test objects, despite being evolved to grip only a ball.

\section{Background}

\subsection{Granular jamming in soft robotics}
Granular jamming is a highly capable means of achieving stiffness-tuneable actuation and gripping of arbitrarily shaped objects by soft robotic devices \cite{fitzgerald_review_2020}. Granular jamming occupies a very useful niche in soft actuation, being particularly useful in applications that seek to exploit it's rapid response time ($\approx$1s) and very large potential stiffness variation  \cite{shintake2018soft}.  

Although recent research has studied a range of different morphologies in jamming actuators, including  worm-like and snake-like robots~\cite{steltz2010jamming,Robertsoneaan6357}, fingers and hands \cite{wei2016novel}, and locomoting spheres \cite{steltz2009jsel}, the predominant jamming actuator is the 'bag' style universal gripper originally described in 2010 \cite{brown_universal_2010} that can be most simply realised using a latex balloon membrane and coffee grounds as the granular material.

Bag grippers have subsequently appeared across a wide range of application domains, including prosthesis \cite{cheng2016prosthetic}, underwater manipulation for deep sea missions \cite{licht2018partially}, industrial gripping \cite{amend2016soft}, and as paws for legged robots \cite{chopra_granular_2020}.  A range of further studies have explored the effects of, e.g., pressure, grain shape, and grain size, using a bag-style gripper to conduct those experiments (e.g., \cite{howard2021shape}).  Bag grippers have also been combined with learning algorithms \cite{jiang2012learning} for object-specific grasp strategies.  Bag style grippers can therefore be seen as a benchmark for jamming actuators. An important question to ask is therefore 'how optimal is a bag shape?'.  This is the question we seek to answer in this paper.

\subsection{Jamming gripper design exploration}

Granular jamming grippers present a range of design variables that can be tuned to elicit specific performance regimes.  Grain size, shape, and softness have been shown to have a significant effect on gripper performance, confirmed by studies of natural \cite{cavallo2019soft}, manufactured \cite{ruotolo2019load} and 3D printed \cite{santarossa2021soft,howard2021shape} grains.  Grains are relatively simple to study as a range of candidates is readily available.  

The same cannot be said for membrane morphology, although studies show that membrane material significantly affects gripper performance \cite{jiang2014robotic}. Preliminary studies have patterned 'nubs' on the inside of the membrane, although their effect on performance was mixed \cite{kapadia_design_2012}
Other preliminary work demonstrates novel multi-material membranes that can induce programmed deformations and improve gripping performance on hard-to-grip objects such as coins \cite{howard_one-shot_2021}.  Membrane morphology is a relatively underexplored area of research as creating a variety of morphologies is only practical through techniques including 3D printing, and despite recent breakthroughs (e.g., \cite{aktacs2021modeling}), modelling the complex interactions between a membrane, the grains, and the environment is an unsolved problem.  This means pure experimental design experimentation based on 3D printing is an attractive option, to circumvent the reality gap and exploit the parallel nature of the fabrication process.

\subsection{Evolutionary robotics}

Evolution has been demonstrated to be a useful enabling technique for bespoke design of granular materials with desired properties optimised for broad applications including soft robotics \cite{jaeger2015celebrating,delaney_multi-objective_2020, delaney_utilising_2019}. Evolution has been applied in-silico to the optimisation of the granular material within a soft robotic gripper \cite{fitzgerald_evolving_2021}, where bespoke grains from a large set of possible morphologies were explored in order to optimise the grip strength, with a complex morphological dependency found on both the size and shape of the target object. These approaches were entirely modelled with no physical instantiation.

Ideally, modelling approaches can be parallelised on a computing cluster as a means to scalability whereby an entire generation of candidate designs can be assessed at once \cite{collins2021review}.  FEA is a typical approach, however it's drawback is the limited range of environmental interactions it can capture.  Mass-spring methods have shown promise in evolving soft voxel robots, however the resulting real robots suffered from some reality gap effects \cite{kriegman2020scalable}.  For a summary of available modelling/design techniques see \cite{pinskier2021bioinspiration}.  

Our approach circumvents these issues and diretly evolves in hardware.   We parallelise on a print bed rather than a high performance computer. Similar approaches using rigid modular robots have been seen before \cite{brodbeck2015morphological}, however we harness printing rather reassembly of fixed modules to explore the design space.

\section{Methodology}
 
 We follow an experimental loop of {\em print, assess, evolve}.  The current generation of $N$=5 grippers are converted from their numerical 'genome' representation into CAD models and printed.   After printing, the grippers have their support removed and are cleaned.  Grippers are then filled with coffee grounds and tested.  Testing provides the average retention (gripping) force on a target object, which is used as a fitness metric.  Fitness is automatically fed back into a desktop computer that acts as an experiment manager, and a genetic algorithm selects and mutates parent genomes to create new children in the next generation.  The experimental loop repeats for a maximum of $G$=15 generations.

 \begin{figure*}[t!]
\centering
 \includegraphics[width=2.0\columnwidth]{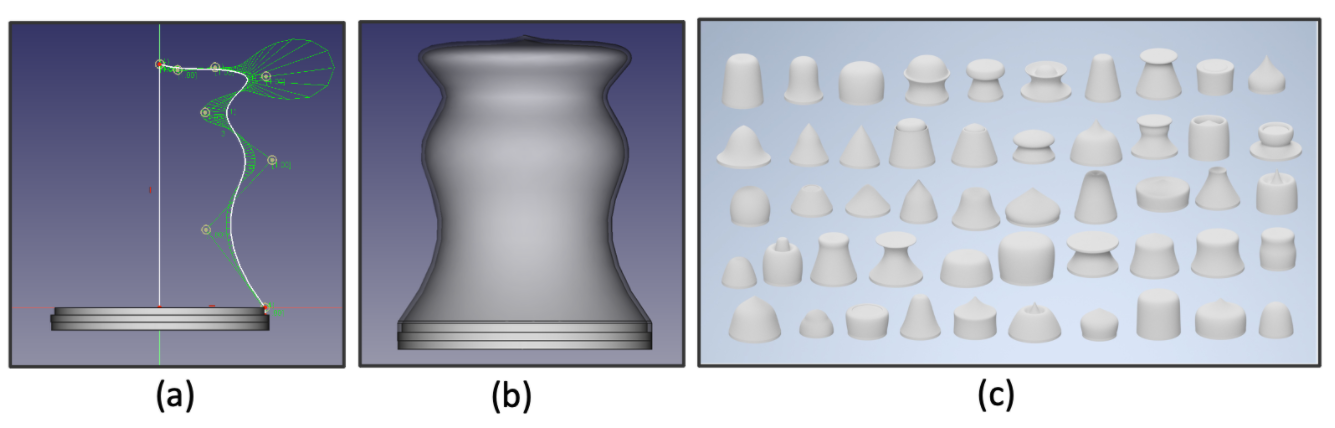}
\caption{Showing the step-by-step process that turns each genotype into a CAD model ready for printing: (a) The genome is a collection of Bezier curves with a variable number of control points.  The length and width of the gripper are included in the genome and fix the first and last control points, as well as setting the radius of the attached base.  (b) The profile created by the Bezier curve is revolved 360$^o$ around the grippers $y$ axis, creating a solid mesh.  (c) A variety of attainable geometries, illustrating the potential scope of design space exploration.}
\label{fig2}
\end{figure*}

 \subsection{Representation}
 Each gripper is directly represented by a 2D Bezier curve; a representation that has previously been shown to be amenable for use in evolving robot components \cite{10.1145/3205455.3205541}.  Each Bezier has a variable number $v$ of control points $c_{0 (x,y)}$...$c_{(v-1)(x,y)}$, plus the radius of the base $r$ and total gripper height $h$  (both mm).  Control points are constrained such that $0\leq c_{(x)},c_{(y)} \leq 1$.  Any self-intersecting curves are randomly reinitialised as they generate invalid meshes.  The first and last control point are constant and not encoded.  They are set to (1,0) and (0,1), representing the contact with the base and contact with the gripper's virtual central axis respectively (see Fig.\ref{fig2}(a)).  
 
 To create a gripper, the genome is fed into FreeCAD via a script, and is transformed from the (0,1) space such that the maximum value of $x$ = $r$ and the maximum value of $y$ = $h$; e.g., the membrane contacts the base at ($r$,0), and contacts the central axis at (0,$h$).  The control points in the genome are subsequently mapped to their real locations using this transform.  A base of appropriate size is automatically generated in CAD as a ring with thickness 1mm, outer radius $r$, and inner radius $15$mm.  A seal is added on top of the base to connect to the membrane. 
 
 The curve defining the membrane morphology is rotated 360$^o$ around the gripper's central axis to create a membrane attached to the base and symmetric around the $y$ axis at $x=0$ (Fig.\ref{fig2}(b)). The membrane is given a wall thickness of 1mm (the minimum size permitted on our 3D printer).  Fig.\ref{fig2}(c) shows 50 randomly initialised grippers, illustrating a wide variety of printable morphologies.  In the initial population, grippers are random-uniformly generated within the parameter ranges presented in Table \ref{table1}.

 \begin{table}[t!]
    \caption{  Gripper genome parameters and ranges.}
    \centering
    \begin{tabular}{ccc}
    \toprule
       Parameter  & Meaning   & Range \\
    \midrule
            $r$     &   Outer radius of gripper base        &   25-40mm \\ 
          $h$     &   Gripper height at central axis      &   30-60mm \\
           $v$     &   Number of Bezier control points    &      2-6 \\  
           $c(x)$      & Control point $x$ component       & 0-1\\
            $c(y)$      & Control point $y$ component      & 0-1\\
    \bottomrule
    \end{tabular}
    \label{table1}
\end{table}
 

\subsection{Printing}
The current generation of grippers is then printed on our Connex3 Objet 500 multimaterial polyjet printer. The membrane is printed in Agilus30 (Shore-A 30) and the base printed in rigid shore-D Vero material.  The seal between the membrane and adaptor is printed in Shore-A 30 with a thin Shore-A 85 blend of Agilus30 and Vero between membrane and adaptor to increase bond strength.  An entire generation of 5 grippers is printed at once in approximately 4 fours; on our printer we can easily scale each generation to a theoretical maximum of 60-80 grippers at once, depending on base size. 

After printing, excess support material was manually removed using a craft knife and water jet.  Each gripper was then filled with coffee grounds using a funnel, repeatedly tapping the membrane to ensure a complete, even fill,  and filling up to the gripper base level.

\subsection{Testing}
\begin{figure}[t!]
\centering
 \includegraphics[width=0.85\columnwidth]{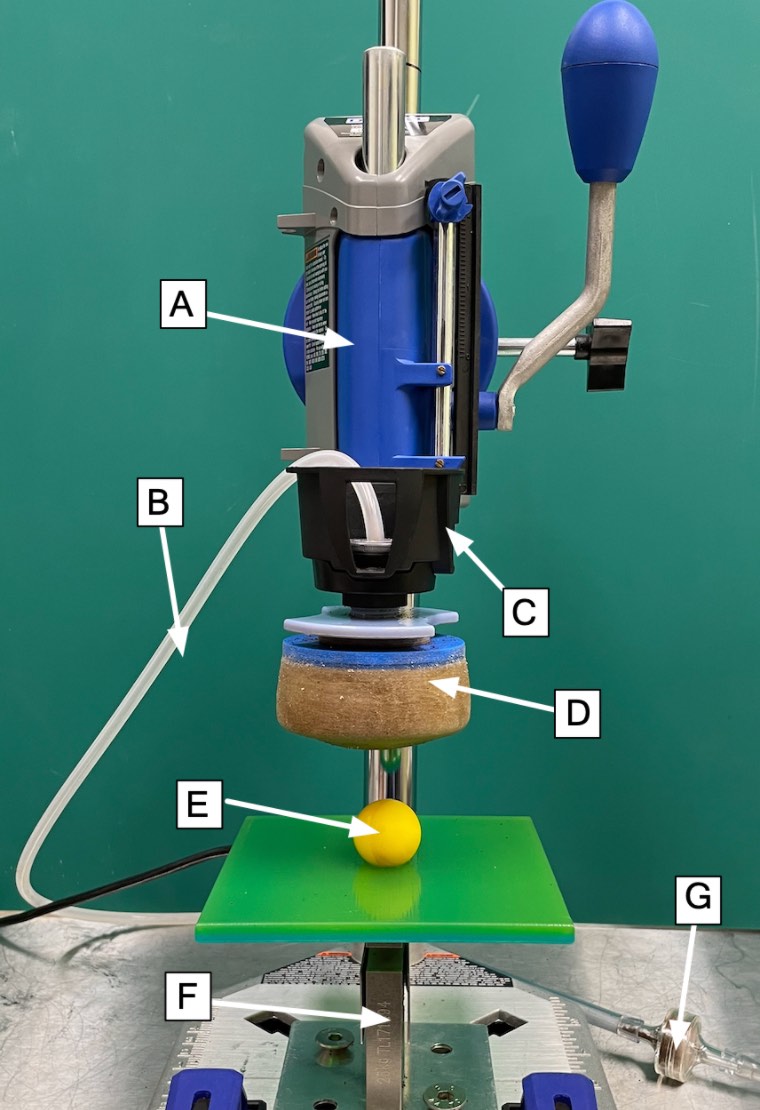}
\caption{Benchtop test setup: A: horizontal drill press, B: silicone tubing (positive and negative pressure) attached to vacuum pump (not shown), C: 3D printed adaptor, D: evolved printed membrane filled with coffee grounds, E: Test object and flat platform, F: load cell, G: vacuum filter.  Data is sent via a USB-serial connection to the desktop PC (not shown).}
\label{testsetup}
\end{figure}

Each gripper was screwed on to a 3D printed adaptor (item C in Fig. \ref{testsetup}), which secures it during testing.  The adaptor connects via M6 silicone tubing and a small filter to a Thomas 107CDC20 H vacuum pump, which provided negative pressure through an SMC IRV10-C06BG regulator at -50 kPa.  The same tubing also connects to an auxiliary positive pressure pump which unjammed the gripper after each test.

The adaptor mounts onto a Dremel linear drill press, which was adjusted so that the tip of the gripper was sitting approximately 30mm above a 25mm radius 3D printed ball test object. The ball is 3D printed with a thread and screwed into a Zemic H3-C3 load cell, which is clamped and centered inline with the  gripper. The load cell records retention force and sends to the desktop PC via a Raspberry Pi 3 Model B+ and a Sparkfun HX711 load cell amplifier.  A 50mm x 50mm platform was attached between the object and load cell to replicate the action of picking an object from a flat surface.  See Fig. \ref{testsetup} for reference.

To test, the gripper is fully lowered onto the test object in an unjammed state.  Negative pressure is applied via the vacuum pump to jam the gripper, which is then slowly raised until it completely releases and clears the test object.  The peak retention force of the grip is recorded and then the vacuum is released. Five one-second bursts of positive pressure are then applied with a 3 second spacing via the auxiliary pump to reset the gripper.  To generate reliable fitness information, each test is repeated five times, and the mean retention force used as the gripper's fitness score.

Once five test results are stored on the raspberry Pi, they are returned to the desktop PC.  Each gripper's fitness score $f$ is set to the mean peak retention force of the five tests.

\subsection{Genetic Algorithm}
 
The gripper design space is explored using a GA, which is a gradient-free black box optimiser suited for multi-modal problem spaces such as ours.  The GA can alter a gripper's radius, height, and can modify the placement of control points, as well as adding or removing control points from the curve. 
  
After all grippers in a generation have their fitness assessed and recorded, the next generation of five children is created via fitness-proportional selection, which gives preference to high-performing parents and is used to balance design space exploration and performance optimisation given the small number of generations \cite{miller2014evolution}.  One parent is always selected in this way; a second parent is chosen in an identical manner if the crossover probability $\chi$=0.8 is satisfied\footnote{All GA probabilities are sampled from a uniform distribution in the range 0-1. Mutation rates were selected following a brief parameter sweep.}.  In this case, one-point crossover is applied such that control points are preserved (i.e., not split by crossover) and that at least one control point is taken from the second parent.  The first peturbation of crossed parents (that with parent 1's genome contribution first) is used and the second peturbation is discarded.  
  
Mutation is applied to each possible allelle in the genome.  For $r$, $h$, plus the two components of each control point $c_x$ and $c_y$, mutation occurs on satisfaction of $\mu$=0.2.   Mutations alter the value of the allelle by an amount drawn from a normal distribution with a standard deviation set to 10\% of the parameter's range.   With probability $\eta$=0.25, either a new control point is randomly initialised and added to the genome (50\% chance), or a random control point is deleted (50\% chance).  All operations respect the bounds in Table \ref{table1}.
 
\begin{figure}[t!]
\centering
\includegraphics[width=0.95\columnwidth]{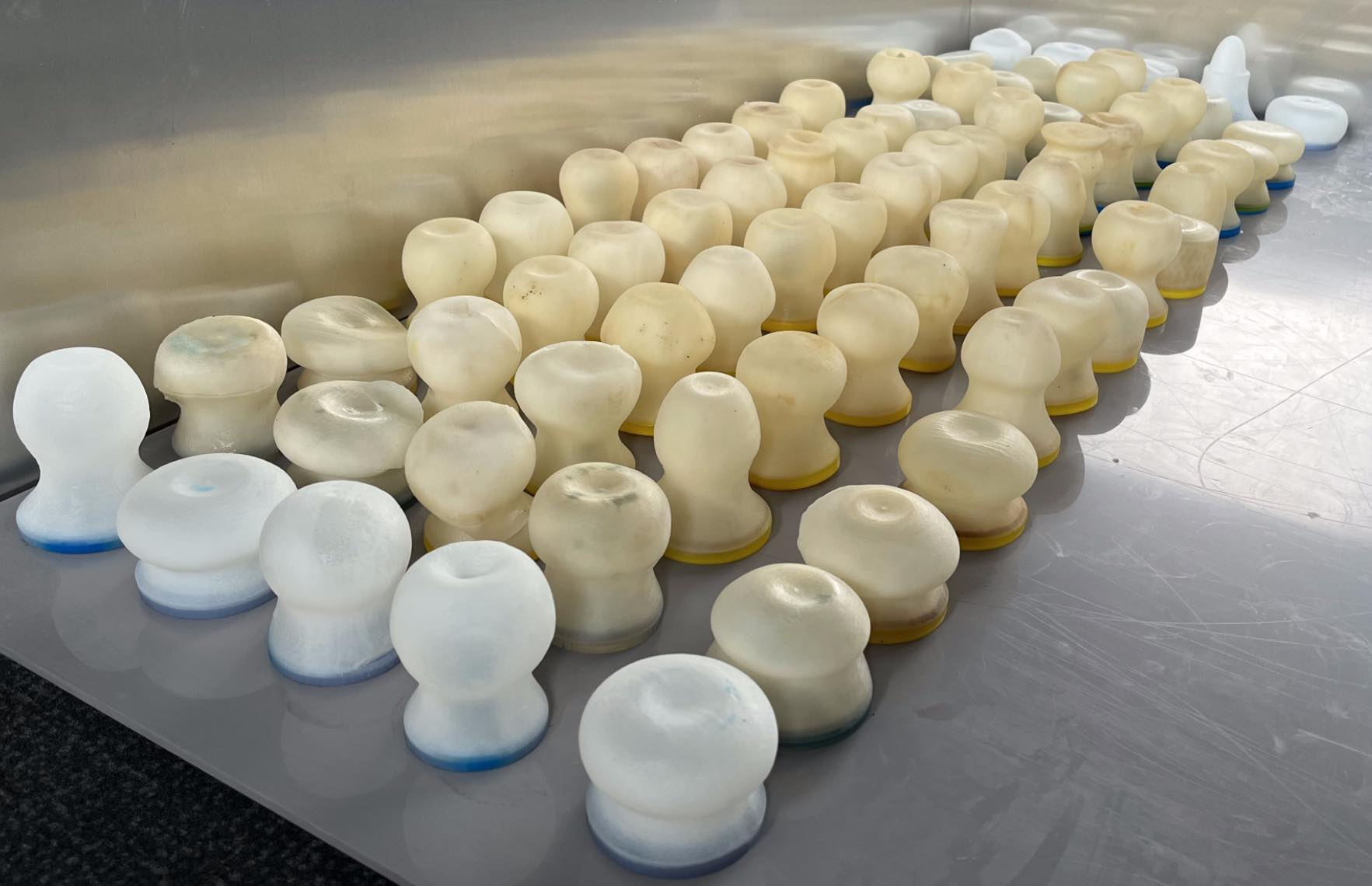} 
\caption{All 75 grippers arranged in generational order, with generation 15 at the front and generation 1 at the rear of the image.  Some grippers were reprinted for the image.}
\label{all-printed}
\end{figure}

The child grippers then printed and tested as before.  Once all grippers have fitness values, the generation is over and a new generation begins.  We run 15 generations of {\em in materio} evolution with 5 grippers per generation, for a total of 75 printed and tested grippers (see Fig.\ref{all-printed}).  With 5 repeats per test, this constitutes a total of 375 real data points, which is used to generate the graphs below.

 \section{Results}
 
 Results assess the evolution of grippers in three main ways, (i) gripper performance, (ii) gripper morphology, and (iii) progression of the evolutionary process.  To create a reasonable baseline, we compare to a 3D printed bag-style gripper.

 \subsection{Performance}
 
\begin{figure}[t!]
\centering
\subfigure{\includegraphics[width=0.95\columnwidth]{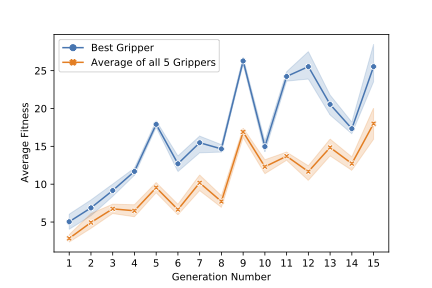} }\\(a)\\
\subfigure{\includegraphics[width=0.95\columnwidth]{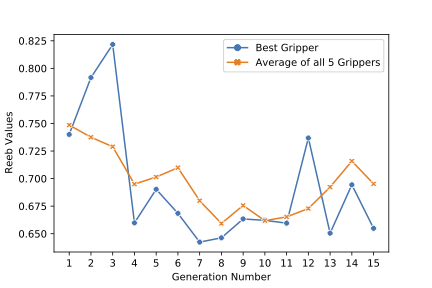} }\\(b)\\
\caption{Evolution of gripper performance through the 15 generations: (a) Maximum retention force (best gripper per generation, average of 5 tests), and Average retention force (mean of 5 tests per gripper, averaged over the entire population). (b) Morphology comparison to bag-style jamming gripper using multiresolution Reeb graphs: best gripper and average similarity per generation.  Higher numbers equate to grippers that are more similar to a bag-style spherical gripper. }
\label{fitness_complexity}
\end{figure}

Fig. \ref{fitness_complexity}(a) shows strong fitness progression through the generations for both maximum and population fitness, with the best gripper per generation starting around 5N and ending at 29N.  The baseline standard bag gripper achieved an average of 2.68N on the same object.

 To investigate how the grippers achieved such strong performance, we note that the best-performing grippers evolve to exploit all three of the mechanisms by which granular grippers generate grip strength \cite{amend_positive_2012}.  {\em Static friction from surface contact} is maximised through evolution of a curves surface that approximates the curve of the sphere to maximise the contact area.  {\em Geometric constraints from interlocking} are evidenced as grippers partially envelope the lower half of the ball when the gripper is pushed over the object.  {\em  Vacuum suction from an airtight seal} is seen in the highest-performing grippers as a deep pocket in the centre of the gripper.  These mechanisms are exploited relatively reliably, as evidenced through the low standard error in Fig. \ref{fitness_complexity}(a).  The discovery of these grippers is facilitated by the Bezier representation which allow for these features to be readily embodied in the grippers.
 
 Features further away from the contact surface serve two main purposes, (i) to 'set up' specific curves on the contact surface, or (ii) are be optimised to cause deformations that pinch around the object when pushed down onto it.

\begin{figure}[t!]
\centering
\subfigure{\includegraphics[width=0.8\columnwidth]{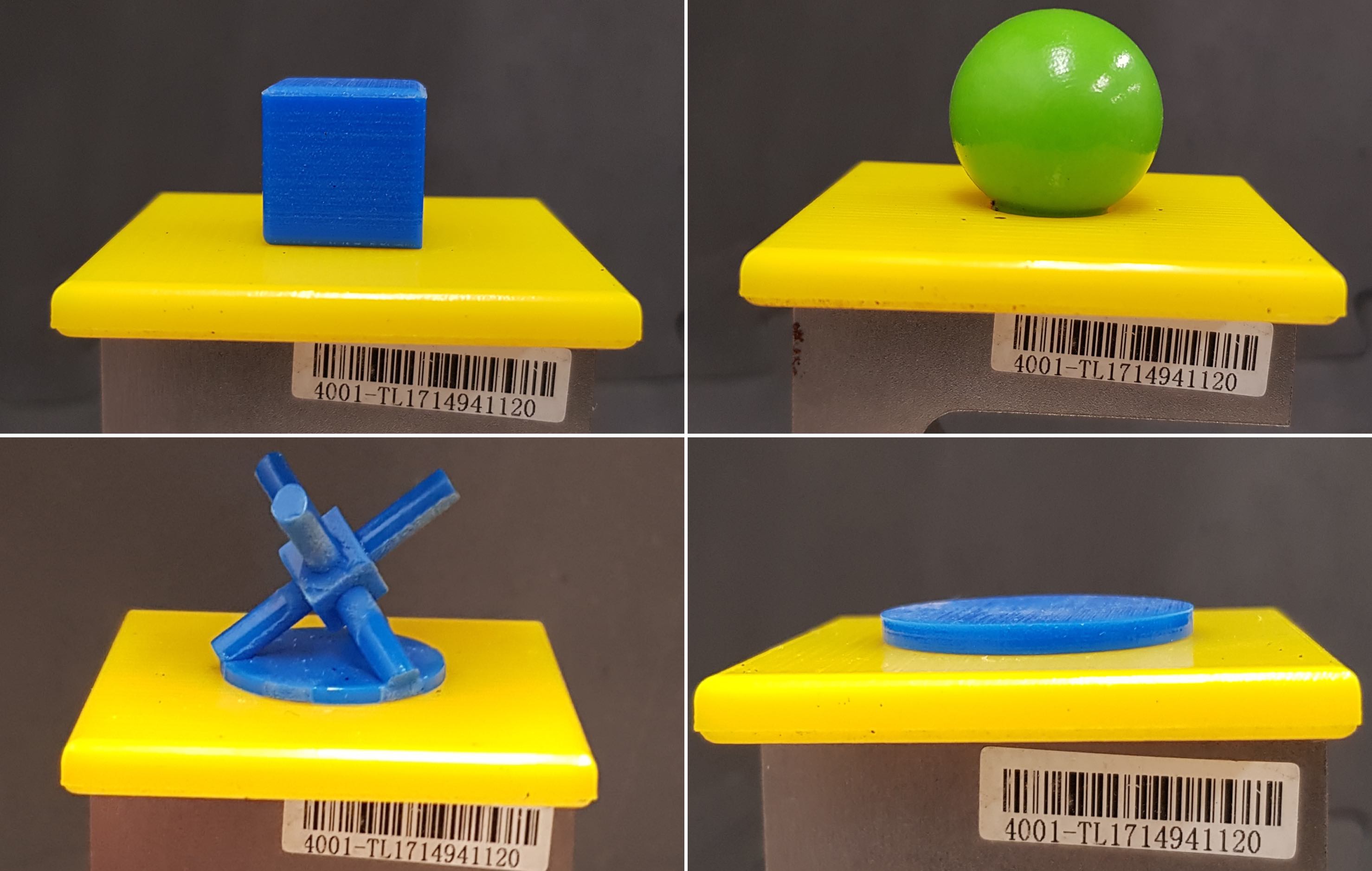} }\\(a)\\
\subfigure{\includegraphics[width=0.99\columnwidth]{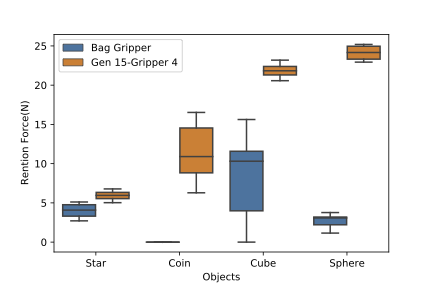} }\\(b)\\
\caption{(a) The four test objects, clockwise from top-left; Cube, Ball, Coin, and Star.(b) Comparing performance (retention force) for a standard bag gripper and the best evolved gripper (generation 15 gripper 4) across a range of test objects. }
\label{fitness_4objects}
\end{figure}

 One potential issue with this approach is overspecialisation caused by testing on a single test object.  To assess the generality of the grippers, we test both the baseline and best evolved gripper on three other challenging and geometrically diverse test objects; a a cube, a star, and a coin with approximate size of 25mm \cite{howard_one-shot_2021} (Fig. \ref{fitness_4objects}(a)).

Fitness, averaged over 5 tests, shows that the evolved grippers transfer well onto these objects (Fig. \ref{fitness_4objects}(b)).  We suggest that transferability is due to the promotion of fundamental gripping mechanisms rather than object-specific features.  The strongest optimisation effect is seen for the Ball object that the grippers were optimised for, however both cube and coin objects show vastly improved performance by the best evolved gripper.  Interestingly, the bag-style gripper shows large average error on the cube; previous studies have highlighted this issue with 3D printed membranes struggling to comply around the shear surfaces.  The evolved gripper appears to solve this issue despite still using a printed membrane.  The star object shows closer (although still higher) performance between the two grippers.  

These findings suggest that results might be more widely applicable, however the limits of this generality are out of scope for this paper and subject to further study.

 \subsection{Morphology}
 
\begin{figure}[t!]
\centering
\includegraphics[width=0.95\columnwidth]{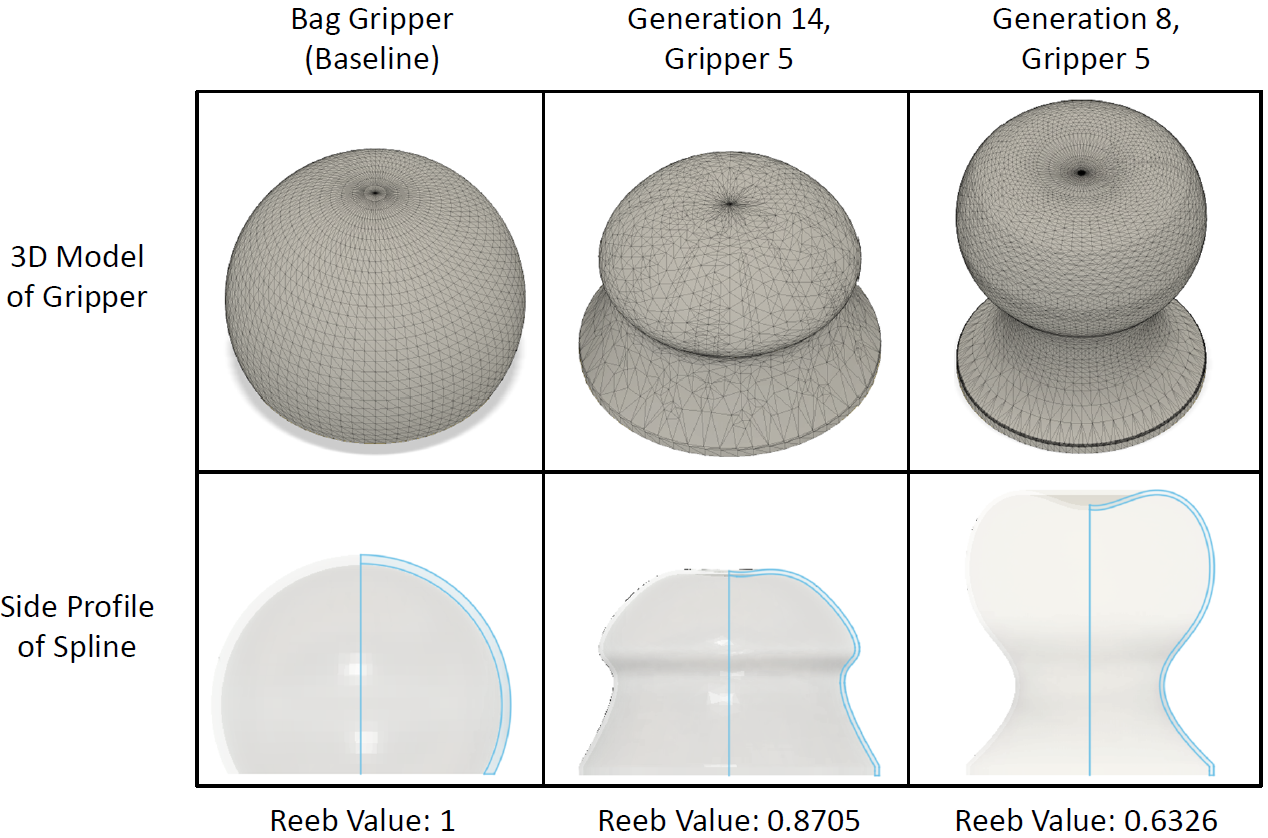} 
\caption{Illustrating the Reeb similarity values attained when comparing evolved grippers to the standard spherical bag gripper.  Higher numbers show increased similarity to a bag gripper, and similarity scores range between 0 and 1.}
\label{GripperComplexityDiagram}
\end{figure}

 Multi-level Reeb graphs \cite{bespalov2003reeb} allow us to assess the morphological similarity of the evolved grippers.  We compare the CAD of each gripper to a spherical bag gripper to generate a similarity score, which ranges between 0 and 1 and where higher values correspond to more similar geometries.  For added context, Fig. \ref{GripperComplexityDiagram} shows example similarity scores.
 
 Fig. \ref{fitness_complexity}(b) shows that similarity generally decreases throughout the generations, e.g., the population (which is initially random) tends to move into a search space away from bag-style geometries.  The best gripper in 12 of the 15 generations displays a similarity score lower than the population average, meaning that fitter grippers are less spherical.  The single best gripper had a similarity score of 0.663.  Similarity scores are limited by the representation, as Beziers have a strong tendency to produce curved surfaces.

 \subsection{Evolutionary process}
 
\begin{figure*}[t!]
\centering
\includegraphics[width=1.5\columnwidth]{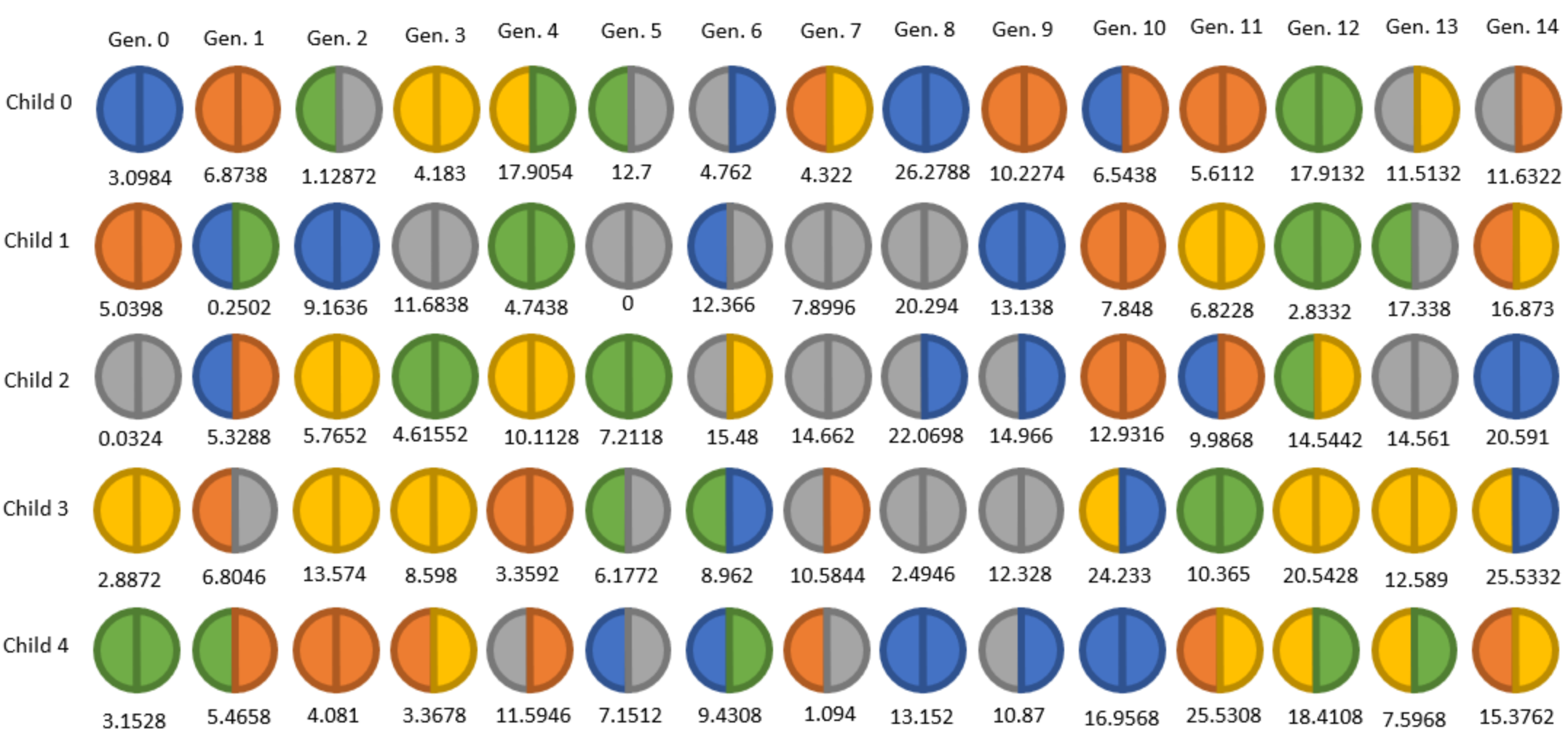} 
\caption{Tracking the heredity of the grippers through the 15 generations.  Individuals may have two colours as two parents may contribute to their genetic code via crossover.  The colours in generation 0 are used throughout to denote lineage, e.g., child 0 in generation 2 results from child 4 and child 2 in the previous generation.  Colours are consistent with the position of the child, e.g., blue always denotes child 0, orange always denotes child 1, etc.   Fitness is shown underneath the gripper.}
\label{lineage}
\end{figure*}

Fig. \ref{lineage} allows us to visualise the evolutionary process, showing the composition of each generation with respect to the parents they are created from, and demonstrating the effects of fitness-proportional selection.  We note that child 1 in generation 5 was given a fitness of 0; this design was not printable.  A low failure rate of 1/75 shows the benefit of the Bezier representation in creating printable grippers, however guaranteed printability is the eventual aim.

 \section{Discussion}
 
This paper showed the first successful model-free direct evolution of soft robotic grippers.  Grippers were evolved over 15 generations to maximise their retention force on a challenging test object.  Results show a strong optimisation effect, with an upward trend in both maximum and average generational gripping forces.  Evolution generated high-performing grippers significantly in excess of the grip performance of a comparative bag-style gripper.  Evolution exploited all three mechanisms delineated in the literature for increasing granular grip strength \cite{amend_positive_2012}, meaning grippers worked well when tested on an assortment of other test objects and suggests that the evolved designs are potentially highly generalisable, although further experimentation is required to confirm.  Overall, direct evolution is demonstrated to be a viable technique for design of soft robotic mechanisms despite operating with relatively low population sizes and few generations.  Evolution works especially well when coupled with 3D printing, as (i) entire generations of (in this case 5, but up to 80) grippers may be printed at once, and (ii) diverse and performant designs can be simply fabricated.  
 
Reeb complexity assessed morphological differences between the evolved grippers and a standard spherical bag gripper, showing increased generational deviation from bag-like geometries as the evolution progressed.  The best gripper per generation was generally less similar to a bag-like gripper than the average similarity of the generation, showing that the best grippers were typically found away from the standard designs used in the literature. This result suggests that, despite their popularity in the literature,  bag-style grippers are frequently not optimal and that other designs should be considered.
 
The ability to design, fabricate and test soft devices at scale is a key component of upcoming 'autonomous design' \cite{pinskier2021bioinspiration} or 'high throughput' methodologies \cite{howard2019evolving,howison2020reality} that generate predictive models through large-scale experimental data collection to further scale design exploration.  Our platform can be seen as a small-scale prototype demonstrator of the data collection part of this methodology, although significant human input is still required to communicate with the printer, clean the prints, and set them up on the experimental apparatus.

 Besides increasing the level of automation in our technique, the most straightforward extension is to expose a wider range of design variables to evolution: grain shape and size, as well as material composition of the membrane and potential surface patterning are all potential optimisation candidates.

\section*{ACKNOWLEDGEMENT}
This work was funded by CSIRO's Active Integrated Matter Future Science Platform; project TB04-WP20 'Designing Jamming Actuation'.


\bibliographystyle{IEEEtran}
\bibliography{jamming}

\begin{thebibliography}{10}
\providecommand{\url}[1]{#1}
\csname url@rmstyle\endcsname
\providecommand{\newblock}{\relax}
\providecommand{\bibinfo}[2]{#2}
\providecommand\BIBentrySTDinterwordspacing{\spaceskip=0pt\relax}
\providecommand\BIBentryALTinterwordstretchfactor{4}
\providecommand\BIBentryALTinterwordspacing{\spaceskip=\fontdimen2\font plus
\BIBentryALTinterwordstretchfactor\fontdimen3\font minus
  \fontdimen4\font\relax}
\providecommand\BIBforeignlanguage[2]{{%
\expandafter\ifx\csname l@#1\endcsname\relax
\typeout{** WARNING: IEEEtran.bst: No hyphenation pattern has been}%
\typeout{** loaded for the language `#1'. Using the pattern for}%
\typeout{** the default language instead.}%
\else
\language=\csname l@#1\endcsname
\fi
#2}}

\bibitem{fitzgerald_review_2020}
\BIBentryALTinterwordspacing
S.~G. Fitzgerald, G.~W. Delaney, and D.~Howard, ``\BIBforeignlanguage{en}{A
  {Review} of {Jamming} {Actuation} in {Soft} {Robotics}},''
  \emph{\BIBforeignlanguage{en}{Actuators}}, vol.~9, no.~4, p. 104, Dec. 2020,
  number: 4 Publisher: Multidisciplinary Digital Publishing Institute.
  [Online]. Available: \url{https://www.mdpi.com/2076-0825/9/4/104}
\BIBentrySTDinterwordspacing

\bibitem{fitzgerald_evolving_2021}
\BIBentryALTinterwordspacing
S.~G. Fitzgerald, G.~W. Delaney, D.~Howard, and F.~Maire, ``Evolving soft
  robotic jamming grippers,'' in \emph{Proceedings of the {Genetic} and
  {Evolutionary} {Computation} {Conference}}, ser. {GECCO} '21.\hskip 1em plus
  0.5em minus 0.4em\relax New York, NY, USA: Association for Computing
  Machinery, June 2021, pp. 102--110. [Online]. Available:
  \url{https://doi.org/10.1145/3449639.3459331}
\BIBentrySTDinterwordspacing

\bibitem{howard_one-shot_2021}
\BIBentryALTinterwordspacing
G.~D. Howard, J.~Brett, J.~O'Connor, J.~Letchford, and G.~W. Delaney,
  ``One-{Shot} {3D}-{Printed} {Multimaterial} {Soft} {Robotic} {Jamming}
  {Grippers},'' \emph{Soft Robotics}, June 2021, publisher: Mary Ann Liebert,
  Inc., publishers. [Online]. Available:
  \url{https://www-liebertpub-com.ezproxy.library.uq.edu.au/doi/full/10.1089/soro.2020.0154}
\BIBentrySTDinterwordspacing

\bibitem{greenwood2006introduction}
G.~W. Greenwood and A.~M. Tyrrell, \emph{Introduction to evolvable hardware: a
  practical guide for designing self-adaptive systems}.\hskip 1em plus 0.5em
  minus 0.4em\relax John Wiley \& Sons, 2006, vol.~5.

\bibitem{shintake2018soft}
J.~Shintake, V.~Cacucciolo, D.~Floreano, and H.~Shea, ``Soft robotic
  grippers,'' \emph{Advanced Materials}, vol.~30, no.~29, p. 1707035, 2018.

\bibitem{steltz2010jamming}
E.~Steltz, A.~Mozeika, J.~Rembisz, N.~Corson, and H.~Jaeger, ``Jamming as an
  enabling technology for soft robotics,'' in \emph{Electroactive Polymer
  Actuators and Devices (EAPAD) 2010}, vol. 7642.\hskip 1em plus 0.5em minus
  0.4em\relax International Society for Optics and Photonics, 2010, p. 764225.

\bibitem{Robertsoneaan6357}
\BIBentryALTinterwordspacing
M.~A. Robertson and J.~Paik, ``New soft robots really suck: Vacuum-powered
  systems empower diverse capabilities,'' \emph{Science Robotics}, vol.~2,
  no.~9, 2017. [Online]. Available:
  \url{https://robotics.sciencemag.org/content/2/9/eaan6357}
\BIBentrySTDinterwordspacing

\bibitem{wei2016novel}
Y.~Wei, Y.~Chen, T.~Ren, Q.~Chen, C.~Yan, Y.~Yang, and Y.~Li, ``A novel,
  variable stiffness robotic gripper based on integrated soft actuating and
  particle jamming,'' \emph{Soft Robotics}, vol.~3, no.~3, pp. 134--143, 2016.

\bibitem{steltz2009jsel}
E.~Steltz, A.~Mozeika, N.~Rodenberg, E.~Brown, and H.~M. Jaeger, ``Jsel:
  Jamming skin enabled locomotion,'' in \emph{2009 IEEE/RSJ International
  Conference on Intelligent Robots and Systems}.\hskip 1em plus 0.5em minus
  0.4em\relax IEEE, 2009, pp. 5672--5677.

\bibitem{brown_universal_2010}
\BIBentryALTinterwordspacing
E.~Brown, N.~Rodenberg, J.~Amend, A.~Mozeika, E.~Steltz, M.~R. Zakin,
  H.~Lipson, and H.~M. Jaeger, ``\BIBforeignlanguage{en}{Universal robotic
  gripper based on the jamming of granular material},''
  \emph{\BIBforeignlanguage{en}{Proceedings of the National Academy of
  Sciences}}, vol. 107, no.~44, pp. 18\,809--18\,814, Nov. 2010, publisher:
  National Academy of Sciences Section: Physical Sciences. [Online]. Available:
  \url{https://www.pnas.org/content/107/44/18809}
\BIBentrySTDinterwordspacing

\bibitem{cheng2016prosthetic}
N.~Cheng, J.~Amend, T.~Farrell, D.~Latour, C.~Martinez, J.~Johansson,
  A.~McNicoll, M.~Wartenberg, S.~Naseef, W.~Hanson, \emph{et~al.}, ``Prosthetic
  jamming terminal device: A case study of untethered soft robotics,''
  \emph{Soft robotics}, vol.~3, no.~4, pp. 205--212, 2016.

\bibitem{licht2018partially}
S.~Licht, E.~Collins, G.~Badlissi, and D.~Rizzo, ``A partially filled jamming
  gripper for underwater recovery of objects resting on soft surfaces,'' in
  \emph{2018 IEEE/RSJ International Conference on Intelligent Robots and
  Systems (IROS)}.\hskip 1em plus 0.5em minus 0.4em\relax IEEE, 2018, pp.
  6461--6468.

\bibitem{amend2016soft}
J.~Amend, N.~Cheng, S.~Fakhouri, and B.~Culley, ``Soft robotics
  commercialization: Jamming grippers from research to product,'' \emph{Soft
  robotics}, vol.~3, no.~4, pp. 213--222, 2016.

\bibitem{chopra_granular_2020}
S.~Chopra, M.~T. Tolley, and N.~Gravish, ``Granular {Jamming} {Feet} {Enable}
  {Improved} {Foot}-{Ground} {Interactions} for {Robot} {Mobility} on
  {Deformable} {Ground},'' \emph{IEEE Robotics and Automation Letters}, vol.~5,
  no.~3, pp. 3975--3981, July 2020, conference Name: IEEE Robotics and
  Automation Letters.

\bibitem{howard2021shape}
D.~Howard, J.~O'Connor, J.~Brett, and G.~W. Delaney, ``Shape, size, and
  fabrication effects in 3d printed granular jamming grippers,'' in \emph{2021
  {IEEE} {International} {Conference} on {Soft} {Robotics} (Robosoft)}, 2021.

\bibitem{jiang2012learning}
Y.~Jiang, J.~R. Amend, H.~Lipson, and A.~Saxena, ``Learning hardware agnostic
  grasps for a universal jamming gripper,'' in \emph{2012 IEEE International
  Conference on Robotics and Automation}.\hskip 1em plus 0.5em minus
  0.4em\relax IEEE, 2012, pp. 2385--2391.

\bibitem{cavallo2019soft}
A.~Cavallo, M.~Brancadoro, S.~Tognarelli, and A.~Menciassi, ``A soft retraction
  system for surgery based on ferromagnetic materials and granular jamming,''
  \emph{Soft robotics}, vol.~6, no.~2, pp. 161--173, 2019.

\bibitem{ruotolo2019load}
W.~Ruotolo, F.~S. Roig, and M.~R. Cutkosky, ``Load-sharing in soft and spiny
  paws for a large climbing robot,'' \emph{IEEE Robotics and Automation
  Letters}, vol.~4, no.~2, pp. 1439--1446, 2019.

\bibitem{santarossa2021soft}
A.~Santarossa, H.~G{\"o}tz, A.~Sack, T.~P{\"o}schel, and P.~M{\"u}ller, ``Soft
  particles reinforce robotic grippers: Robotic grippers based on granular
  jamming of soft particles,'' \emph{arXiv preprint arXiv:2109.03356}, 2021.

\bibitem{jiang2014robotic}
A.~Jiang, T.~Ranzani, G.~Gerboni, L.~Lekstutyte, K.~Althoefer, P.~Dasgupta, and
  T.~Nanayakkara, ``Robotic granular jamming: Does the membrane matter?''
  \emph{Soft Robotics}, vol.~1, no.~3, pp. 192--201, 2014.

\bibitem{kapadia_design_2012}
J.~Kapadia and M.~Yim, ``Design and performance of nubbed fluidizing jamming
  grippers,'' in \emph{2012 {IEEE} {International} {Conference} on {Robotics}
  and {Automation}}, May 2012, pp. 5301--5306, iSSN: 1050-4729.

\bibitem{aktacs2021modeling}
B.~Akta{\c{s}}, Y.~S. Narang, N.~Vasios, K.~Bertoldi, and R.~D. Howe, ``A
  modeling framework for jamming structures,'' \emph{Advanced Functional
  Materials}, vol.~31, no.~16, p. 2007554, 2021.

\bibitem{jaeger2015celebrating}
H.~M. Jaeger, ``Celebrating soft matter’s 10th anniversary: Toward jamming by
  design,'' \emph{Soft matter}, vol.~11, no.~1, pp. 12--27, 2015.

\bibitem{delaney_multi-objective_2020}
\BIBentryALTinterwordspacing
G.~W. Delaney and G.~Howard, ``Multi-objective exploration of a granular matter
  design space,'' in \emph{Proceedings of the 2020 {Genetic} and {Evolutionary}
  {Computation} {Conference} {Companion}}, ser. {GECCO} '20.\hskip 1em plus
  0.5em minus 0.4em\relax New York, NY, USA: Association for Computing
  Machinery, July 2020, pp. 263--264. [Online]. Available:
  \url{https://doi.org/10.1145/3377929.3389951}
\BIBentrySTDinterwordspacing

\bibitem{delaney_utilising_2019}
G.~W. Delaney, D.~Howard, and K.~D. Napoli, ``Utilising {Evolutionary}
  {Algorithms} to {Design} {Granular} {Materials} for {Industrial}
  {Applications},'' in \emph{2019 18th {IEEE} {International} {Conference} {On}
  {Machine} {Learning} {And} {Applications} ({ICMLA})}, Dec. 2019, pp.
  1897--1902.

\bibitem{collins2021review}
J.~Collins, S.~Chand, A.~Vanderkop, and D.~Howard, ``A review of physics
  simulators for robotic applications,'' \emph{IEEE Access}, 2021.

\bibitem{kriegman2020scalable}
S.~Kriegman, A.~M. Nasab, D.~Shah, H.~Steele, G.~Branin, M.~Levin, J.~Bongard,
  and R.~Kramer-Bottiglio, ``Scalable sim-to-real transfer of soft robot
  designs,'' in \emph{2020 3rd IEEE International Conference on Soft Robotics
  (RoboSoft)}.\hskip 1em plus 0.5em minus 0.4em\relax IEEE, 2020, pp. 359--366.

\bibitem{pinskier2021bioinspiration}
J.~Pinskier and D.~Howard, ``From bioinspiration to computer generation:
  Developments in autonomous soft robot design,'' \emph{Advanced Intelligent
  Systems}, p. 2100086, 2021.

\bibitem{brodbeck2015morphological}
L.~Brodbeck, S.~Hauser, and F.~Iida, ``Morphological evolution of physical
  robots through model-free phenotype development,'' \emph{PloS one}, vol.~10,
  no.~6, p. e0128444, 2015.

\bibitem{10.1145/3205455.3205541}
\BIBentryALTinterwordspacing
J.~Collins, W.~Geles, D.~Howard, and F.~Maire, ``Towards the targeted
  environment-specific evolution of robot components,'' in \emph{Proceedings of
  the Genetic and Evolutionary Computation Conference}, ser. GECCO '18.\hskip
  1em plus 0.5em minus 0.4em\relax New York, NY, USA: Association for Computing
  Machinery, 2018, p. 61–68. [Online]. Available:
  \url{https://doi.org/10.1145/3205455.3205541}
\BIBentrySTDinterwordspacing

\bibitem{miller2014evolution}
J.~F. Miller, S.~L. Harding, and G.~Tufte, ``Evolution-in-materio: evolving
  computation in materials,'' \emph{Evolutionary Intelligence}, vol.~7, no.~1,
  pp. 49--67, 2014.

\bibitem{amend_positive_2012}
J.~R. Amend, E.~Brown, N.~Rodenberg, H.~M. Jaeger, and H.~Lipson, ``A
  {Positive} {Pressure} {Universal} {Gripper} {Based} on the {Jamming} of
  {Granular} {Material},'' \emph{IEEE Transactions on Robotics}, vol.~28,
  no.~2, pp. 341--350, Apr. 2012.

\bibitem{bespalov2003reeb}
D.~Bespalov, W.~C. Regli, and A.~Shokoufandeh, ``Reeb graph based shape
  retrieval for cad,'' in \emph{International Design Engineering Technical
  Conferences and Computers and Information in Engineering Conference}, vol.
  36991, 2003, pp. 229--238.

\bibitem{howard2019evolving}
D.~Howard, A.~E. Eiben, D.~F. Kennedy, J.-B. Mouret, P.~Valencia, and
  D.~Winkler, ``Evolving embodied intelligence from materials to machines,''
  \emph{Nature Machine Intelligence}, vol.~1, no.~1, pp. 12--19, 2019.

\bibitem{howison2020reality}
T.~Howison, S.~Hauser, J.~Hughes, and F.~Iida, ``Reality-assisted evolution of
  soft robots through large-scale physical experimentation: a review,''
  \emph{Artificial Life}, vol.~26, no.~4, pp. 484--506, 2020.

\end{thebibliography}
\balance

\end{document}